\documentclass[11pt,a4paper]{article}
\usepackage[hyperref]{acl2020}
\usepackage{times}
\usepackage{comment}
\usepackage{mathtools}
\usepackage{amssymb}
\usepackage{placeins}
\usepackage{latexsym}
\usepackage{multirow}

\usepackage{graphicx}
\usepackage{microtype}

\aclfinalcopy 

\title{Pushing on Personality Detection from Verbal Behavior:\newline A Transformer Meets Text Contours of Psycholinguistic Features}
\author{Elma Kerz\\
  RWTH-Aachen University\\
  \texttt{\normalsize elma.kerz@ifaar.rwth-aachen.de} \\\And
  Yu Qiao \\
  RWTH-Aachen University\\
  \texttt{\normalsize yu.qiao@rwth-aachen.de} \\
  \AND
  Sourabh Zanwar\\
  RWTH-Aachen University\\
  \texttt{\normalsize sourabh.zanwar@rwth-aachen.de}\\
  \And
  Daniel Wiechmann\\
  University of Amsterdam\\
  \texttt{\normalsize d.wiechmann@uva.nl} \\
 }

\date{}

\begin{document}
\maketitle
\begin{abstract}
Research at the intersection of personality psychology, computer science, and linguistics has recently focused increasingly on modeling and predicting personality from language use. We report two major improvements in predicting personality traits from text data: (1) to our knowledge, the most comprehensive set of theory-based psycholinguistic features and (2) hybrid models that integrate a pre-trained Transformer Language Model BERT and Bidirectional Long Short-Term Memory (BLSTM) networks trained on within-text distributions (`text contours') of psycholinguistic features. We experiment with BLSTM models (with and without Attention) and with two techniques for applying pre-trained language representations from the transformer model - `feature-based' and `fine-tuning'.  We evaluate the performance of the models we built on two benchmark datasets that target the two dominant theoretical models of personality: the Big Five Essay dataset \cite{pennebaker1999linguistic} and the MBTI Kaggle dataset \cite{li2018feature}. Our results are encouraging as our models outperform existing work on the same datasets. More specifically, our models achieve improvement in classification accuracy by 2.9\% on the Essay dataset and 8.28\% on the Kaggle MBTI dataset.  In addition, we perform ablation experiments to quantify the impact of different categories of psycholinguistic features in the respective personality prediction models.

\end{abstract}

\section{Introduction}

Personality is broadly defined as the combination of a person's behavior, emotions, motivation, and characteristics of thought patterns \cite{corr2020cambridge}. Our personality has a major impact on our lives, influencing our life choices, well-being, health, and preferences and desires \cite{ozer2006personality}. Specifically, personality has been repeatedly linked to individual (e.g., happiness, physical and mental health), interpersonal (e.g., quality of relationships with peers, family, and romantic partners), and social-institutional outcomes (e.g., career choice, satisfaction and achievement, social engagement, political ideology) \cite{soto2019replicable}. 

While there are several models of human personality, the predominant and widely accepted model is the Big Five or Five Factor Model \cite{mccrae1992introduction,mccrae2009five}. In this model, personality traits are divided into five factors: (1) Extraversion (assertive, energetic, outgoing, etc.), (2) Agreeableness (appreciative, generous, compassionate, etc.), (3) Conscientiousness (efficient, organized, responsible, etc.), (4) Neuroticism (anxious, self-pitying, worried, etc.), and (5) Openness (curious, empathetic, imaginative, etc.). These five personality traits are commonly assessed by questionnaires in which a person reflects on his or her typical patterns of thinking and behavior, such as the NEO Five Factor Inventory \cite{costa1992neo}, and the Big-Five Inventory \cite{john1991big}; \citep[see][for a comprehensive overview]{matthews2009}.  The Myers–Briggs Type Indicator (MBTI) is another widely administered questionnaire, in particular in applied settings \cite{meyers1990introduction}. In contrast to the Big Five personality taxonomy, which conceptualizes human personality as latent trait scores, the MBTI model describes personality in terms of 16 types that result from combining binary categories into four dimensions: (a) Extraversion/Introversion (E/I) - preference for how people direct and receive their energy, based on the external or internal world, (b) Sensing/Intuition (S/N) - preference for how people take in information, through the five senses or through interpretation and meanings, (c) Thinking/Feeling (T/F) - preference for how people make decisions, relying on logic or emotion over people and particular circumstances, and (d) Judgment/Perception (J/P) - how people deal with the world, by ordering it or remaining open to new information.

Given its central importance in capturing the essential aspects of human life, increasing attention is being paid to the development of models that can leverage behavioral data to automatically predict personality.  Data obtained from verbal behavior is one of the key types of such data. Even in the early years of psychology, a person's use of language was seen as a distillation of his or her underlying drives, emotions, and thought patterns \cite[see][for historical overviews]{tausczik2010psychological,boyd2017language}. 
Early approaches to automatic personality prediction (APP) -- also referred to as automatic personality prediction or recognition -- from textual data have relied on machine learning models based on psycholinguistic features, whereas more recent approaches to APP typically draw on deep learning techniques that use pre-trained word embeddings \cite[see][for an overview of the former]{vinciarelli2014survey} \cite[see][for an overview of deep learning-based APP]{mehta2020recent}. 

In this paper, we make a valuable contribution to this dynamic area of APP research by presenting two important improvements in predicting personality traits from textual data: (1) to our knowledge, the most comprehensive set of psycholinguistic features and (2) hybrid models that integrate a pre-trained Transformer Language Model BERT and Bidirectional Long Short-Term Memory (BLSTM) networks trained on in-text distributions ('text contours') of psycholinguistic features. Since our goal is to demonstrate the utility of our modeling approach, we conduct our experiments on two widely used benchmark datasets: the Big Five Essay dataset \cite{pennebaker1999linguistic} and the MBTI-Kaggle dataset \cite{li2018feature}, which align with the dominant personality models described above.  
The remainder of this paper is organized as follows: In Section 2, we briefly review recent related work on these two benchmark datasets. Then, in Section 3, we present the two benchmark datasets and the extraction of psycholinguistic features using automated text analysis based on a sliding window approach. In Section 4, we describe our modeling approach, and in Section 5, we present and discuss the results. Finally, we conclude with possible directions for future work in Section 6.



\section{Related work}

\citet{majumder2017deep} used a convolutional neural network (CNN) feature extractor in which sentences were fed to convolution filters to obtain n-gram feature vectors. Each individual text of the  Big Five Essay dataset was represented by aggregating the vectors of its sentences and the obtained vectors were concatenated with psycholinguistic (Mairesse) features \cite{mairesse2007using}. For  classification, they fed the resulting document vector to a fully connected neural network with one hidden layer. Using this method, they were able to achieve an average classification accuracy of 58\% for the Big Five personality traits on the Essays dataset. \citet{kazameini2020personality} were the first to use a Transformer-Based Language model to extract contextualized word embeddings. Specifically, they built a Bagged-SVM classifier fed with contextualized embeddings extracted from BERT, a pre-trained language model with a Bidirectional Encoder from Transformers \cite{devlin2018bert}. Their model outperformed the CNN-based model proposed by the \citet{majumder2017deep} model by 1.04\%.  \citet{amirhosseini2020machine} used a Gradient Boosting Model (GBM) based on Term Frequency–Inverse-Document-Frequency features (TF-IDF)  to predict personality dimensions in the Kaggle MBTI dataset. Their modeling approach achieved an average classification accuracy across all dimensions of 76.1\%. Using both the Big Five Essay dataset and the Myers-Briggs’ type indicator Kaggle Dataset, \citet{mehta2020bottom} proposed the integration of deep learning models and psycholinguistic features with language model embeddings for APP. They extracted a total of 123 psycholinguistic features, including the Mairesse features set \cite{mairesse2007using}, SenticNet \citep{cambria2010senticnet}, NRC-Emotion Lexicon \citep{mohammad2013crowdsourcing}, and NRC-VAD Lexicon \citep{mohammad2018obtaining}.
Language model features were extracted using BERT. Their experiments compared the performance of BERT-base and BERT-large in synergy with SVM or Multi-layer Perceptron (MLP) classifiers. BERT-base + MLP yielded an average score of 60.6 on the Essay dataset, while BERTlarge + MLP yielded an average score of 77.1 on the Kaggle dataset.  The approach taken in \citet{mehta2020bottom} outperformed the previously best-performing model by \citet{amirhosseini2020machine} by 1\%.
Zooming on classification accuracy for specific personality traits, the models in \newcite{mehta2020bottom} achieved the highest performance on two of the Big Five personality traits in the Essays dataset (openness, accuracy = 64.6\%, and conscientiousness, accuracy = 59.2\%) and on three of the four MBTI dimensions in the Kaggle MBTI dataset (Intuitive/Sensing (N/S), accuracy = 86.6\%, Thinking/Feeling (T/F), accuracy = 76.1\% and Perception/Judging (P/J), accuracy = 67.2\%).  The highest performance on the Introversion/Extraversion (I/E) MBTI dimension (79\%) was obtained by the `GBM + TFIDF' model reported in \newcite{amirhosseini2020machine}. The highest performance on the three remaining Big Five dimensions was achieved recently by \newcite{ramezani2021automatic}, which used an ensemble modeling approach (stacking) to combine linguistic and ontology-based features with deep learning-based methods based on a hierarchical attention network as a meta-model. Although the overall performance of SOTA on the Essay dataset was not superior - mainly due to relatively poor performance on the Openness trait (accuracy = 56.3\%), this work has demonstrated the utility of model stacking as an effective way to boost the prediction of personality traits. For a performance overview of the models reviewed here for different data sets and personality dimensions, see Table \ref{tab:classification} in Section 4.

\section{Method}

\subsection{Datasets}

We conducted our experiments with two widely used personality benchmark datasets: (1) The Essays Dataset \citep{pennebaker1999linguistic} and (2) Kaggle MBTI Dataset \citep{li2018feature}. (1) Essays: This stream-of-consciousness dataset consists of 2468 essays written by students and annotated with the binary labels of the Big Five personality traits, which were obtained through a standardized self-report questionnaire. The average text length is 672 words and the total size of the dataset is approximately 1.6 million words.  One of the reasons why Essays is an established benchmark dataset is the relatively large amount of continuous language use and the fact that the personality traits were obtained using a validated instrument. (2) Kaggle MBTI: This dataset was collected through the PersonalityCafe forum\footnote{\url{https://www.personalitycafe.com/}} and thus provides a diverse sample of people interacting in an informal online social environment. It consists of samples of social media interactions from 8675 users, all of whom indicated their MBTI type.  The average text length is 1,288 words. The total size of the entire dataset is approximately 11.2 million words.


\subsection{Measurement of text contours of psycholinguistic features}

\begin{figure}
    \centering
    \includegraphics[width = 0.5\textwidth]{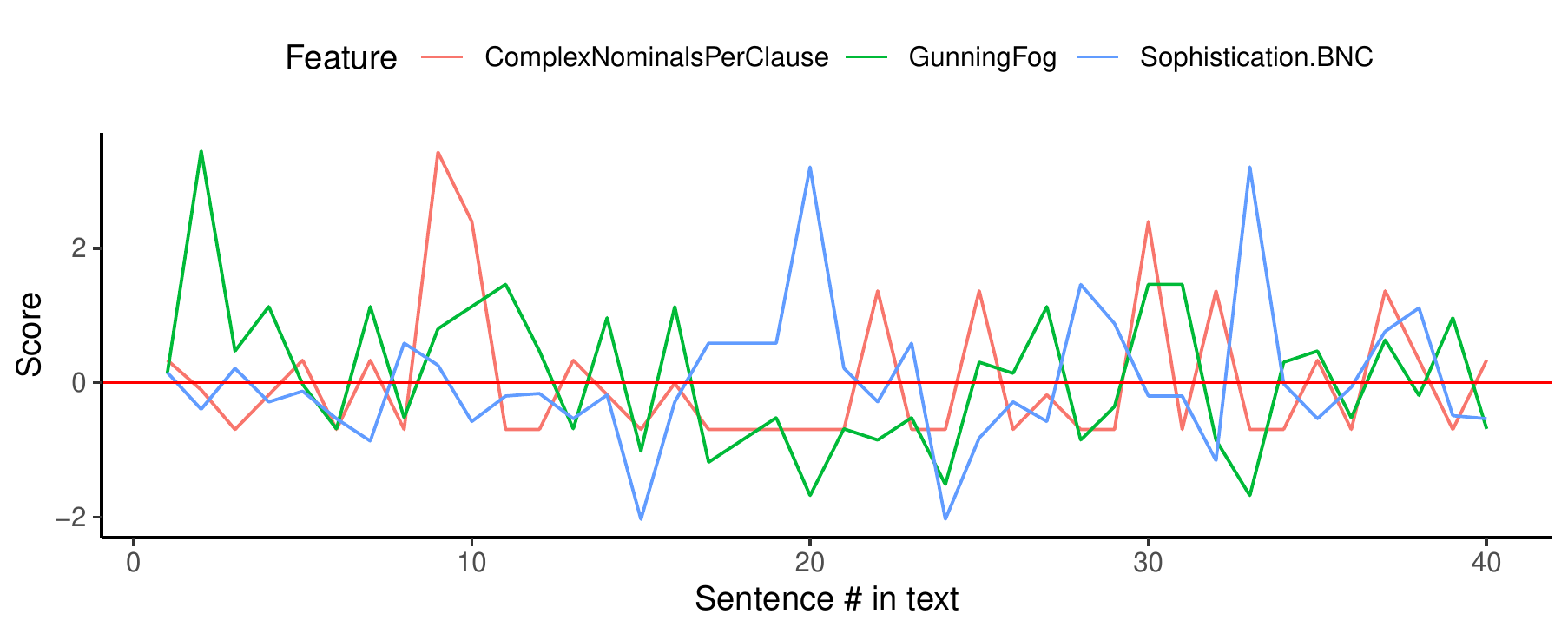}
    \caption{Text contours for three selected features of first 40 sentences of a randomly selected text from the Essays dataset (ID: 2004 499).}
    \label{fig:contours1}
\end{figure}

The texts from both datasets (the Big Five Essay dataset and the MBTI Kaggle dataset) were automatically analyzed using an automated text analysis (ATA) system that employs a sliding window technique to compute sentence-level measurements.  These measurements capture the within-text distributions of scores for a given psycholinguistic feature, referred to here as ‘text contours’ (for recent applications of the ATA system in the context of text classification, see \cite{kerz2020becoming,qiao2021alzheimer,qiao2021prediction}. We extracted a set of 437 theory-based psycholinguistic features that can be binned into four groups: (1) features of morpho-syntactic complexity (N=19), (2) features of lexical richness, diversity and sophistication (N=77), (3) readability features (N=14), and (4) lexicon features designed to detect sentiment, emotion and/or affect (N=326). Tokenization, sentence splitting, part-of-speech tagging, lemmatization and syntactic PCFG parsing were performed using Stanford CoreNLP \citep{manning2014stanford}.
The group of \textbf{morpho-syntactic complexity features} includes (i) surface features related to the length of production units, such as the average length of clauses and sentences, (ii) features of the type and frequency of embeddings, such as number of dependent clauses per T-Unit or verb phrases per sentence and (iii) the frequency of particular structure types, such as the number of complex nominals per clause. This group also includes (iv) information-theoretic features of morphological and syntactic complexity based on the Deflate algorithm \citep{deutsch1996rfc1951}. 
The group of \textbf{lexical richness, diversity and sophistication features} includes six different subtypes: (i) lexical density features, such as the ratio of the number of lexical (as opposed to grammatical) words to the total number of words in a text, (ii) lexical variation, i.e. the range of vocabulary as manifested in language use, captured by text-size corrected type-token ratio, (iii) lexical sophistication, i.e. the proportion of relatively unusual or advanced words in a text, such as the number of words from the New General Service List \citep{browne2013new}, (iv) psycholinguistic norms of words, such as the average age of acquisition of the word \citep{kuperman2012age} and two recently introduced types of features: (v) word prevalence features that capture the number of people who know the word \citep{brysbaert2019word,johns2020estimating} and (vi)  register-based n-gram frequency features that take into account both frequency rank and the number of word n-grams ($n\in [1,5]$). The latter were derived from the five register subcomponents of the Contemporary Corpus of American English \citep[COCA, 560 million words,][]{davies2008corpus}: spoken, magazine, fiction, news and academic language \citep[see][for details see e.g.]{kerz2020becoming}. The group of  \textbf{readability features} combines a word familiarity variable defined by a prespecified vocabulary resource to estimate semantic difficulty along with a syntactic variable, such as average sentence length. Examples of these measures include the Fry index \citep{fry1968readability} or the SMOG \citep{mclaughlin1969clearing}. 
The group of  \textbf{lexicon-based sentiment/emotion/affect features (SentiEmo)} was derived from a total of ten lexicons that have been successfully used in personality detection, emotion recognition and sentiment analysis research: (1) The Affective Norms for English Words (ANEW) \citep{bradley1999affective}, (2) ANEW-Emo lexicons \citep{stevenson2007characterization}, (3) DepecheMood++ \citep{araque2019depechemood++}, (4) The Geneva Affect Label Coder (GALC) \citep{scherer2005emotions}, (5) The General Inquirer \citep{stone1966general}, (6) The LIWC dictionary \citep{pennebaker2001linguistic}, (7) The NRC Word-Emotion Association Lexicon \citep{mohammad2013crowdsourcing}, (8) The NRC Valence, Arousal, and Dominance (NRC-VAD) lexicon \citep{mohammad2018obtaining}, (9) SenticNet \citep{cambria2010senticnet}, and (10) the Sentiment140 lexicon \citep{MohammadKZ2013}. The feature value for each subcategory in a given lexicon is the mean value of all rated/scored words in a given sentence.
The informational gain of `text contours' compared to text-averages is illustrated in Figure \ref{fig:contours1}. The Figure shows the distribution of  z-standardized values of three selected features for a randomly selected text from the Essay dataset. The red line represents the average feature value of the text. As can be seen from the graphs, all feature values fluctuate within the text, with high values for one feature often offset by lower values for another. The contour-based classifiers, discussed in more detail in Section 3, can take advantage of this high-resolution assessment of psycholinguistic features.

\section{Modeling approach}
Our models are constructed from three components: (a) a `contour encoder' that converts a sequence of psycholinguistic features into a hidden representation vector, (b) a pre-trained transformer-based language model, BERT, that converts a sequence of tokens into a hidden representation vector, and (c) a classifier that outputs the probability of a personality feature given the hidden representation of the sample. We conduct experiments with three types of personality prediction models: (1) contour encoder + classifier, (2) hybrid models that combine the contour encoder with a transformer-based language model + classifier, and (3) a stacking model that combines ten repetitions of the best performing model. As for the contour encoder, we experiment with BLSTM and BLSTM with attention models. Attention-based models have been successfully used in a variety of tasks, including machine translation \cite{bahdanau2014neural}, speech recognition \cite{huang2016attention} and relation classification \cite{zhou2016attention}. In the context of personality classification, learning a scoring function gives sentence weighting to the attention mechanism and allows a model to pay more attention to the most influential sentences in a text for a personality trait. As for the hybrid models, we experiment with different strategies for applying the pre-trained language model - `feature-based' and `fine-tuning': In the feature-based approach, we freeze model weights during training and use the pre-trained contextualized word embeddings from BERT. In the `fine-tuning' approach, we unfreeze all 12 layers and fine-tune towards the personality detection task \citep[see][]{devlin2018bert}. 

All models are implemented using PyTorch \citep{paszke2019pytorch}. Unless specifically stated otherwise, we use binary cross entropy as our loss function, 'AdamW' as optimizer, a fixed learning rate of $8\times10^{-4}$ and $dropout=0.1$, $l2=1\times10^{-4}$ as the regularization. The optimal network structures and values of hyperparameters were found by grid-search.
The performance of the models is evaluated by 10-fold cross-validation (ten repetitions) to counter variability due to initialization of the weights. We report the results of the best performing models in comparison to the performance of the APP systems presented in Section 2 Table \ref{tab:classification}. 

\subsection{Components}
\textbf{Contour Encoder:} The contour encoder, $\text{Encoder}_{PSYLING}(X)$,  transforms a sequence of psycholinguistic features $X=(x_1, x_2,\dots, x_n)$ to a hidden psycholinguistic representation vector $P_{PSYLING}$ of a given text. Here, $x_i$ is a 436 dimensional vector for the $i$th sentence obtained from the APA system described in Section 3.2. In this paper, two architectures of contour encoder are applied: BLSTM and BLSTM with attention (ATTN). The BLSTM contour encoder is a $L$-layer BLSTM with number of hidden states of $d_h$. The hidden representation from this model is a $d_{o}=2d_h$ dimensional vector, which is a concatenation of the last hidden states of the last layer in forward ($\overrightarrow{h_n}$) and backward direction ($ \overleftarrow{h_1}$). Specifically, $X \mapsto \text{Encoder}_{BLSTM}(X)=P$:
\begin{equation*}
	\begin{aligned}
	&[\overrightarrow{H}, \overleftarrow{H}] = BLSTM(X) \\
	&P = [\overrightarrow{h_n}^T|\overleftarrow{h_1}^T]^T\\
	\end{aligned}
\end{equation*}
where $[\cdot | \cdot]$ is concatenation operator, $\overrightarrow{H}=(\overrightarrow{h}_1, \overrightarrow{h}_2, \dots, \overrightarrow{h}_n)$ and $\overleftarrow{H}=(\overleftarrow{h}_1, \overleftarrow{h}_2, \dots, \overleftarrow{h}_n)$ are BLSTM model's last layer hidden states in the forward and backward direction. 

The ATTN contour encoder model was constructed as follows: Given a input sequence $X$, a sequence of weights will be computed with the help of a BLSTM model. Then the hidden representation of a given text can be obtained by computing the weighted sum of (a) concatenated hidden vectors from the last layer of the BLSTM model in forward and backward direction (b) feature vectors in $X$. We also experimented with (c) computing weights for each individual dimension of $x_i$ and then taking weighted sum of $X$ by applying this weights. Our experiments shows, that the approach (c) works best for both dataset. So in this paper, we define $X\mapsto \text{Encoder}_{ATTN}$(X)=P:
\begin{equation*}
	\begin{aligned}
	&H = BLSTM(X)\\
	&M = \text{Tanh}(W_{att}H + b_{att}) \\
	&\pmb{\alpha} = \text{Softmax}(M)\\
	&V = \textstyle \sum_{i=1}^n \pmb{\alpha_i}\odot \pmb{x_i}\\
	&P = \text{Tanh}(W_{pool}V + b_{pool})
	\end{aligned}
\end{equation*}
where $W_{att}\in\mathbb{R}^{436\times d_o}, b_{att}\in\mathbb{R}^{436}$. $H$ and $d_o$ is defined as in BLSTM encoder description. Softmax is defined as:
$\alpha_{ij} = \frac{e^{m_{ij}}}{\sum_{k=1}^{n}e^{m_{kj}}}$

\textbf{BERT Language Model:} We use a pre-trained BERT transformer model, `bert-base-uncased', from Huggingface's transformers library \citep{wolf2019huggingface}. The model consists of 12 transformer layers with a hidden size of 768 and 12 attention heads. Texts are tokenized using BERT's BPE tokenizer. We use as input to BERT language model the initial 512 tokens $T = (t_1, t_2, \dots, t_m)$ of a given text, i.e. up to 510 word tokens plus the  [cls] token at the beginning and the  [sep] token at the end of a given text). Assuming the output of the $l$ layer of BERT is $H^{(l)} = (h^{(1)}_1, h^{(l)}_2, \dots, h^{(l)}_n)$, then a hidden vector is computed by either (a) the output for the [cls]-token, i.e. i.e., $V = h^{(l)}_1$ or by (b) averaging the output at the position of the actual tokenized words, i.e., $V = \frac{1}{m - 2}\sum_{i=i}^{m-2}h^{(l)}_i$. Experiments with both approaches for $l\in[1, 12]$ revealed that that (a) the latter approach consistently works better than the former and (b) that $l=11$ works best for the Essays dataset, wheras $l=12$ works best for the MBTI dataset. So we define $X\mapsto\text{Encoder}_{BERT}(T) = P$

\begin{equation*}
	\begin{aligned}
	&H^{(l)} = \text{BERT}(T)\\
	&V = \textstyle \frac{1}{m - 2}\sum_{i=i}^{m-2}h^{(l)}_i\\
	&P = \text{Tanh}(W_{pool}V + b_{pool})\\
	\end{aligned}
\end{equation*}

\textbf{Classifier:} We use a multi-layer feed-forward neural network as our classifier component. The input to the classifier has a dimension corresponding to the underlying encoder's output dimension. We use PReLU as the activation function. Batch normalization was applied between layers of the classifier. All hidden layers share a same hidden size.

\subsection{Models}

 We first construct models based solely on psycholinguistic features.
 These models (1) serve as interpretable baselines for the hybrid prediction models and (2) allow us to determine feature importance of individual features groups in predicting personality traits. To fully utilize the information provided by the contour-based measurement of text features, the models rely on BLSTM or BLSTM with attention architecture, i.e. at position of $\text{Encoder}_{PSYLING}$, $\text{Encoder}_{BLSTM}$ or $\text{Encoder}_{ATTN}$ is applied.

  \begin{equation*}
	\begin{aligned}
	&P =\text{Encoder}_{PSYLING}(X) \\
	&y=\text{Classifier}(P)
	\end{aligned}
\end{equation*}

$\text{Encoder}_{BLSTM}$ has 3 layers with 256 hidden states. We applied a learning rate of 0.001 during training of this model. The BLSTM in $\text{Encoder}_{ATTN}$ has 3 layers with 512 hidden states. The classifier has 3 layers with hidden size of 512.

Our hybrid architecture combines text contours of psycholinguistic features with Transformer-based language models using a late-fusion method by concatenating the hidden representations from the psycholinguistic contour encoder and BERT, specifically
\begin{equation*}
	\begin{aligned}
	&P_{PSYLING} =\text{Encoder}_{PSYLING}(X) \\
	&P_{BERT} = \text{Encoder}_{BERT}(T)\\
	&P = [P_{PSYLING}^T|P_{BERT}^T]^T\\
	&y=\text{Classifier}(P)
	\end{aligned}
\end{equation*}

At the position of $\text{Encoder}_{PSYLING}$, $\text{Encoder}_{BLSTM}$ can be used, which has 3 layers with hidden states of 256, or $\text{Encoder}_{ATTN}$, of which BLSTM also has 3 layers with hidden states of 256 with $dropout=0.2$. During training, parameters of BERT has a fixed learning rate of $2\times 10^{-5}$ while learning rate of $8\times10^{-5}$ is applied to other parameters. The classifier has 3 layers with hidden size of 512.

The final model used in our experiments employed a stacking approach to ensemble our best performing models \citep{wolpert1992stacked}, which has been shown to effectively increase the accuracy of the ensembled individual models. Specifically, we employed model stacking to combine BERT+ATTN-PSYLING (FT) model instances for both dataset. 

The training procedure consists of two stages: In stage one, we take the model prediction on the dev-fold of each model trained on the train-fold of a k-fold CV. These predictions are then concatenated and constitute the one dimension out of 10 of the input data in a subsequent stage (stage 2). We did the same for all 10 iterations. The final predictions of the model are derived from another logistic regression model trained on the concatenated prediction vectors from stage 1 (10-fold CV).

\subsection{Feature importance} 

To assess the relative importance of the feature groups, we employed Submodular Pick Lime (SP-LIME; \newcite{ribeiro2016should}). SPLIME is a method to construct a global explanation of a model by aggregating the weights of linear models, that locally approximate the original model. To this end, we first constructed local explanations using LIME. Analogous to super-pixels for images, we categorized our features into four groups – lexical richness, morphosyntactic complexity, readability, sentiment/emotion (see section 3.2). We used binary vectors $z\in\{0,1\}^{d}$ to denote the absence and presence of feature groups in the perturbed data samples, where $d$ is the number of feature groups. Here, `absent' means that all values of the features in the feature group are set to 0, and `present' means that their values are retained. For simplicity, a linear regression model was chosen as the local explanatory model. An exponential kernel function with Hamming distance and kernel width $\sigma=0.75\sqrt{d}$ was used to assign different weights to each perturbed data sample. After constructing their local explanation for each data sample in the original dataset, the matrix $W\in\mathbb{R}^{n\times d}$ was obtained, where $n$ is the number of data samples in the original dataset and $W_{ij}$ is the $j$th coefficient of the fitted linear regression model to explain data sample $x_i$. The global importance score of the SP-LIME for feature $j$ can then be derived by: $I_j = \sqrt{\sum_{i=1}^n |W_{ij}|}$

\section{Results and Discussion}

\begin{table*}[]
\setlength{\tabcolsep}{2pt}
\caption{Performance comparison (classification accuracy) of our models (bottom) with previous state-of-the-art-models (top). Best performance indicated in bold.}
\begin{tabular}{l|cccccc|cccccl}
\hline
                          & \multicolumn{6}{c}{Essays}                                                                                                                             & \multicolumn{5}{c}{MBTI Kaggle}                                                                                                         &  \\
                        
& O    & C    & E    & A    & N    & Avg  & I/E  & N/S  & T/F  & P/J  & Avg &  \\
Majumder et al. (2018)  & 61.1 & 56.7 & 58.1 & 56.7 & 57.3 & 58   & -    & -    & -    & -    & -       &  \\
Kazameini et al (2020)                      & 62.1 & 57.8 & 59.3 & 56.5 & 59.4 & 59   & -   & -   & - & - & -    &  \\
Amirhosseini \& Kazemian (2020)                      & - & - & - & - & - & -   &79   & 86   & 74.2 & 65.4 & 76.1    &  \\
\hline
\textit{Mehta et al (2020):}   & &&&&&&&&&& \\
Psycholinguistic + MLP    & 60.4 & 57.3 & 56.9 & 57   & 59.8 & 58.3 & 77.6 & 86.3 & 72   & 61.9 & 74.5    &  \\
BERT-base + MLP           &64.6 &59.2 & 60   & 58.8 & 60.5 &60.6 & 78.3 & 86.4 & 74.4 & 64.4 & 75.9    &  \\
All features (base) + MLP & 61.1 & 57.4 & 57.9 & 58.6 & 60.5 & 59.1 & 78.4 &86.6 & 75.9 & 64.4 & 76.3    &  \\
BERT-large + MLP          & 63.4 & 58.9 & 59.2 & 58.3 & 58.9 & 59.7 & 78.8 & 86.3 &76.1 &67.2 &77.1    &  \\
                          \hline
Ramezani et al. (2021) &56.30 &59.18 &\textbf{64.25} &60.31 &61.14& 60.24&-&-&-&-&-\\
\hline
\hline
\textit{Psycholinguistic models (ours)}   & &&&&&&&&&& \\
BLSTM-PSYLING    & 61.69&59.22&58.12&56.87&57.52&58.68&77.29&86.31&72.91&61.01&74.38 \\
ATTN-PSYLING    & 63.15&59.79&59.18&58.29&59.79&60.04&77.29&86.19&73.97&63.69&75.29 \\
\textit{Hybrid models (ours)}   & &&&&&&&&&& \\
BERT+BLSTM-PSYLING (FB)&64.25    &60.80 &60.92&59.26&60.48&61.14&78.39&86.58&74.42&64.17&75.89 \\
BERT+ATTN-PSYLING (FB)& 64.78   & 61.13 &60.44&59.30&60.68&61.27&78.82&86.78&76.62&65.78&77.00 \\
BERT+BLSTM-PSYLING (FT)&65.55    &60.72 &60.72&60.52&\textbf{62.14}&61.93&85.78&90.86&83.79&79.79&85.06 \\
BERT+ATTN-PSYLING (FT)    &66.23 &60.60&61.61&\textbf{61.05}&61.65&62.28&\textbf{86.25}&90.96&84.66&79.65&85.38 \\
BERT+PSYLING Ensemble    &\textbf{71.95} &\textbf{61.38}&63.01&60.16&60.98&\textbf{63.50}&85.47&\textbf{92.27}&\textbf{85.70}&\textbf{82.58}&\textbf{86.51} \\
\hline
\end{tabular}
\label{tab:classification}
\end{table*}

An overview of the results of our models in comparison to those reported in the previous studies reviewed above is presented in Table \ref{tab:classification}.  As Table 1 shows, we achieve state-of-the-art (SOTA) results on both benchmark personality datasets: On the Big Five Essay dataset, our best-performing model achieves a classification accuracy of 63.5\%, which corresponds to an increase  of 2.9\% over the previous SOTA.  On the MBTI Kaggle dataset, our best model improved the classification accuracy of SOTA by 8.28\%. On both datasets the highest classification accuracy was achieved by the ensemble model, which combined ten iterations of a hybrid model integrating a fine-tuned BERT model with an attention-based BLSTM model trained on text contours (see BERT+PSYLING Ensemble in Table 1).  Our models achieve the highest performance on four of the Big Five - all except Extraversion - and on all four MBTI dimensions, with the largest increase in performance for the Big Five on the Openness dimension (+7.35\%) and for the MBTI on the T/F dimension (+9.6\%). Comparing the accuracy for each personality trait from Table \ref{tab:classification} for the hybrid models trained with the "feature-based" strategy (denoted by "FB") with the corresponding value for the models trained with the "fine-tuning" strategy (denoted by "FT"), we find that the accuracy of all traits improved when each pre-trained model was fine-tuned on the data set.  Comparing the accuracy for each personality trait for the models trained with an attention mechanism (denoted by `ATTN') to the corresponding value for the models trained without this mechanism (denoted by `BLSTM'), we find that accuracy on all dimensions except the MBTI N/S improved when an attention mechanism was used. Our results also show that approaches grounded in interpretable features can achieve competitive performance with Transformer-based approaches: Our best-performing model trained solely on psycholinguistic features, the attention-based BLSTM model (ATT-PSYLING), achieved an average classification accuracy of 60.04\%, approaching the previous SOTA model, BERT-base + MLP \newcite{mehta2020bottom}, by only 0.54\%. 
This is a promising finding given the need for more interpretable personality prediction models that can provide valuable insights into key psycholinguistic features to drive personality prediction and advance personality psychology research. See e.g. Rudin (2019) for more general calls for using white-box models to solve practical problems, particularly in the context of critical industries such as healthcare, criminal justice, and news.  This is due to the fact that human experts in a given application domain require both accurate and understandable models (Loyola-Gonzalez, 2019)\nocite{loyola2019black}. 

\begin{table*}
\centering
\caption{Results of the feature ablation for Big Five Essays datset (top) and Kaggle MBTI dataset (bottom): Feature importance (Model: ATTN-PSYLING) macro-averaged across 100 model instances. (10 $\times$ 10-fold CV).}
\scalebox{0.95}{
\begin{tabular}{|c|c|c|c|c|c|c|c|c|c|}
    \hline
    \multicolumn{2}{|c}{O}&\multicolumn{2}{|c}{C}&\multicolumn{2}{|c}{E}&\multicolumn{2}{|c}{A}&\multicolumn{2}{|c|}{N}\\
    \hline
    Group&I&Group&I&Group&I&Group&I&Group&I\\
    \hline
    SentiEmo&18.49&SentiEmo&21.36&SentiEmo&16.39&SentiEmo&9.28&SentiEmo&16.62\\
    lexical&12.90&lexical&14.48&lexical&10.93&lexical&7.52&lexical&10.23\\
    readability&9.57&readability&9.57&morph.syn&9.17&morph.syn&6.23&morph.syn&8.11\\
    morph.syn&7.08&morphosyntactic&8.91&readability&7.51&readability&4.21&readability&7.06\\
    \hline
\end{tabular}
}
\label{tab:FIessays}
\vspace{-3mm}
\begin{tabular}{|c|c|c|c|c|c|c|c|}
    \hline
    \multicolumn{2}{|c}{I/E}&\multicolumn{2}{|c}{N/S}&\multicolumn{2}{|c}{T/F}&\multicolumn{2}{|c|}{P/J}\\
    \hline
    Group&I&Group&I&Group&I&Group&I\\
    \hline
    SentiEmo&33.73&SentiEmo&21.32&SentiEmo&45.06&SentiEmo&24.97\\
    lexical&29.94&lexical&14.25&lexical&24.64&readability&17.21\\
    morph.syn&20.65&readability&12.55&morph.syn&20.31&morph.syn&16.02\\
    readability&18.33&morph.syn&10.40&readability&18.76&lexical&14.48\\
    \hline
\end{tabular}

\label{tab:FI}
\vspace{-3mm}
\end{table*}

In what follows, we present the results of the ablation experiments. Feature group importance was quantified using SP-LIME on the best performing model trained only on text contours of psycholinguistic features, the ATTN-PSYLING model. The results of the feature ablation experiment are presented in Table \ref{tab:FI}. The table shows that the prediction of personality traits was influenced by all four feature groups (all I $>$ 4.21). Overall, personality traits were best predicted by the sentiment/emotion/affect (SentiEmo) feature group. The lexical richness, diversity and sophistication group consistently ranked second on all traits except the P/J MBTI dimension. This result indicates that in addition to words associated with affective-emotional categories, personality traits are also related to more general aspects of vocabulary. Morphosyntactic complexity and readability play a minor role but still achieve high I-scores compared to the highest scoring group in predicting Extraversion, Neuroticism, and Agreeableness (ratio: I(group\textsubscript{j}) / I(SentEmo) $>$ 0.45). Finally, zooming in on the specific interactions between psycholinguistic cues and personality traits, we calculated the difference between the average feature scores of text samples with different labels for each personality trait. Visualizations of the most important psycholinguistic features that influence the prediction of personality traits are shown in Figures \ref{fig:vis1} and \ref{fig:vis2} in the Appendix. Some interesting patterns emerged: For example, texts produced by extroverts tend to (a) have less complex morphosyntax than those by introverts (as indicated by the lower scores of the information-theoretic complexity measures), (b) contain a greater proportion of positive words, and (c) have a higher proportion of frequently used n-grams from the spoken language, news, and magazine registers. The language use of individuals scoring high on Neuroticism showed (a) a higher proportion of self-referencing words, (b) higher proportions of words related to sadness, anxiety and disappointment, but also (c) a higher proportion of longer n-grams from the fiction register. Highly conscientious individuals showed (a) a higher proportion of words with high prevalence, i.e. words that are known by a larger percentage of the population, (b) more words associated with affiliation (ally, friend) and (c) a higher proportions of frequently used n-grams from the academic register. These results replicate and extend previous findings reported in the literature \citep[for overviews see, e.g.,][]{mairesse2007using,park2015automatic,boyd2021natural}.

\section{Conclusion}

Due to its central importance in capturing the essential aspects of human life, increasing attention is being paid to the modeling and predicting personality traits. In this work, we made valuable contributions to advance the state of the art in automatic prediction of personality traits from verbal behavior.  We demonstrated that models trained with a comprehensive set of theory-based psycholinguistic features can compete with a Transformer-based model when their within-text distribution is taken into account.  Moreover, we showed that hybrid models incorporating such features can improve the performance of pre-trained Transformer language models, even when the latter is based on a larger model (BERT-large).  We also showed that different techniques for applying pre-trained language representations from the Transformer model have an impact on model performance.  Our ablation experiments have yielded interesting insights into the interplay between theory-based psycholinguistic features and personality traits. Here, we decided to focus on the two most widely used benchmark datasets. In our future work, we intend to conduct experiments with more recent, larger personality datasets such as PANDORA \citep{matejmladen2020}. Since this dataset also includes metadata (gender, age, and location/region), it would be interesting to see how they contribute to modeling and predicting personality traits from language use.

\FloatBarrier

\bibliography{lrec2022.bib}
\bibliographystyle{acl_natbib}

\appendix

\section{Appendices}

\newpage

\begin{figure}
    \centering
    \includegraphics[width = 0.5\textwidth]{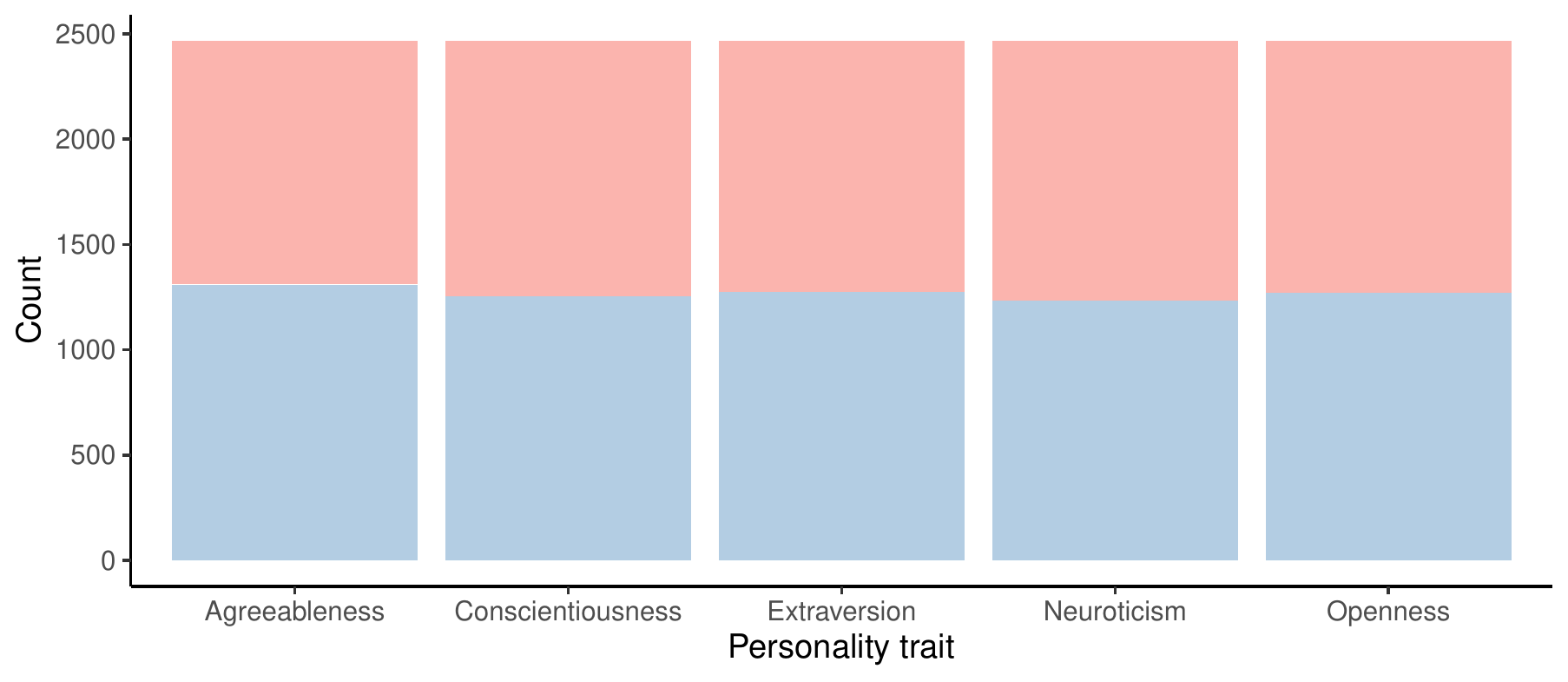}
    \caption{Distribution of labels in the Essay dataset}
    \label{fig:dist1}
\end{figure}

\begin{figure}
    \centering
    \includegraphics[width = 0.5\textwidth]{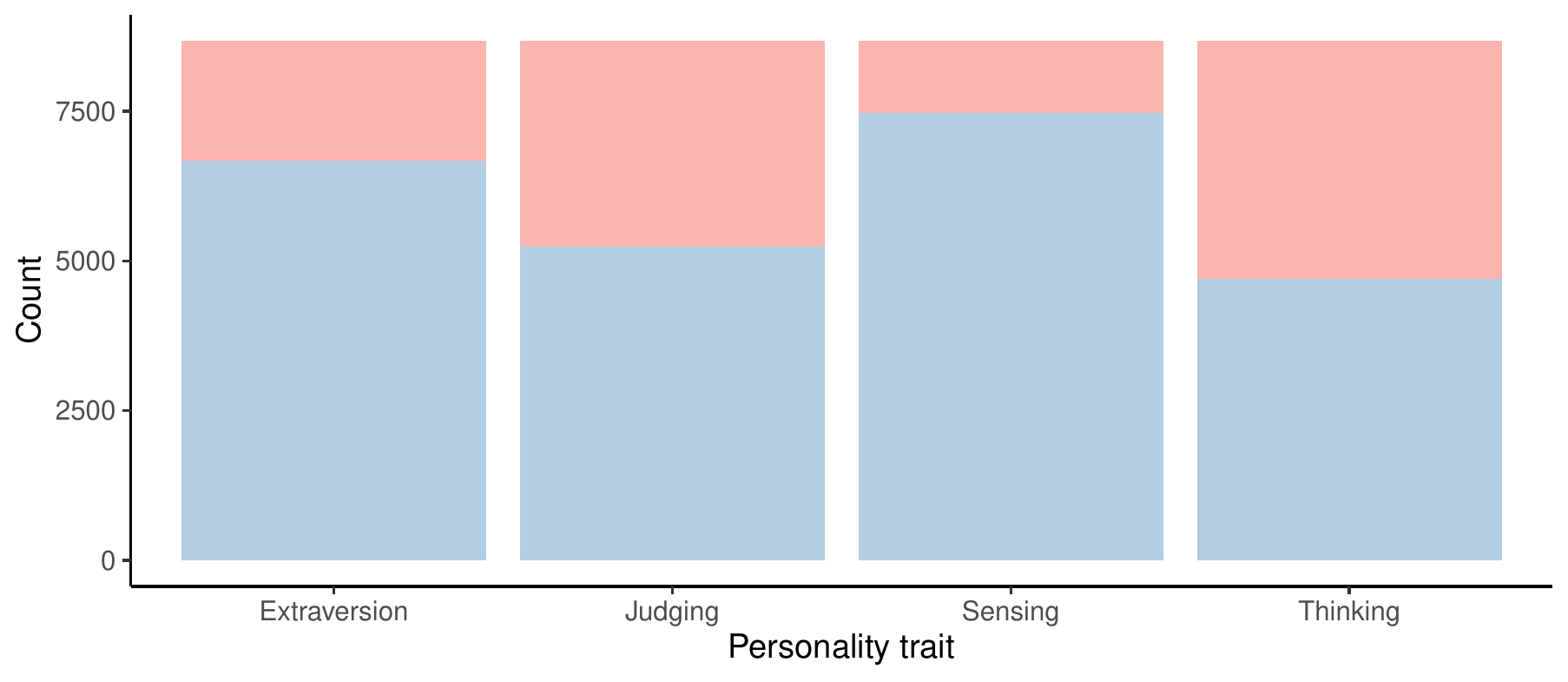}
    \caption{Distribution of labels in the Kaggle MBTI dataset}
    \label{fig:dist2}
\end{figure}

\begin{figure*}
    \centering
    \includegraphics[width = 0.9\textwidth]{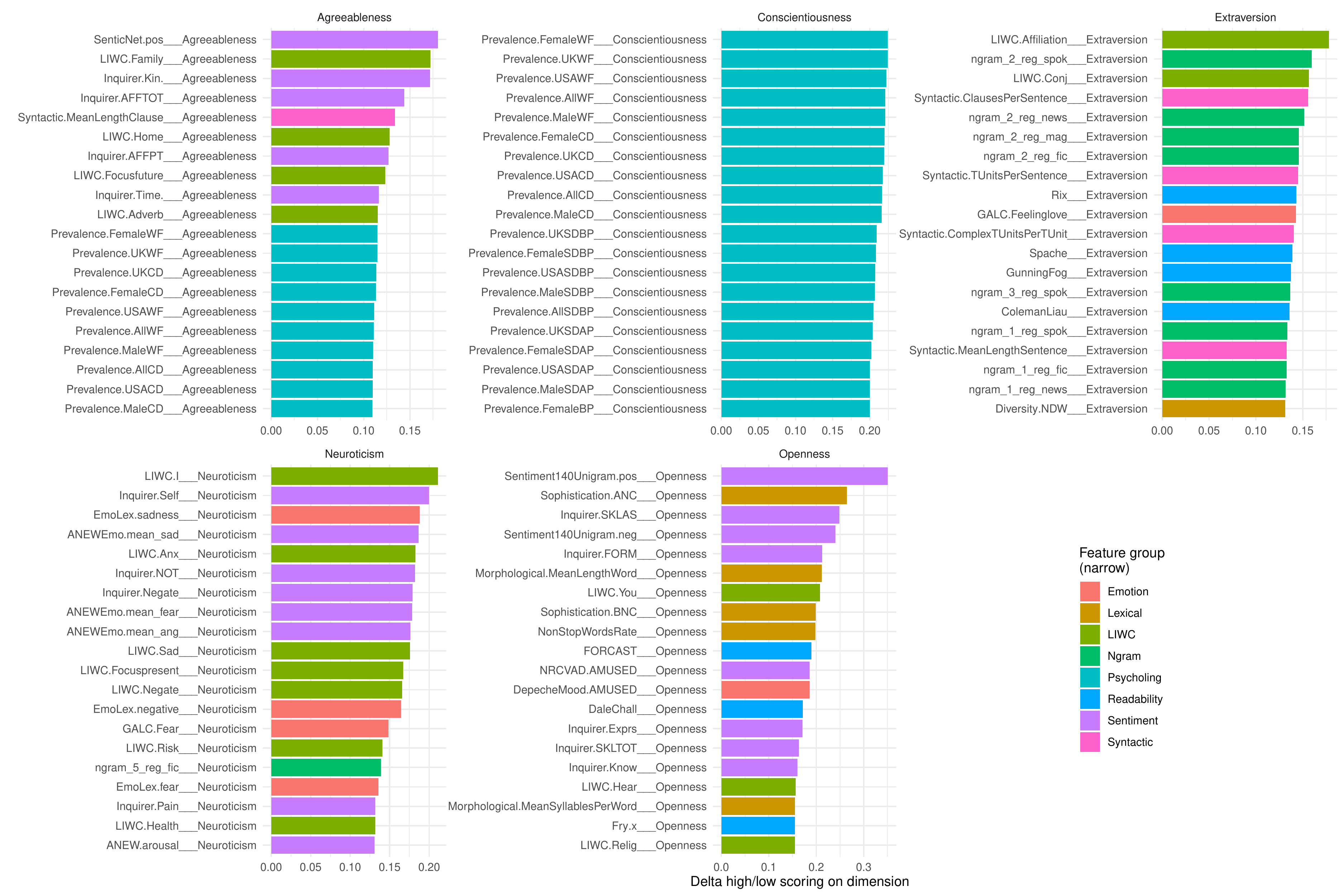}
    \includegraphics[width = 0.9\textwidth]{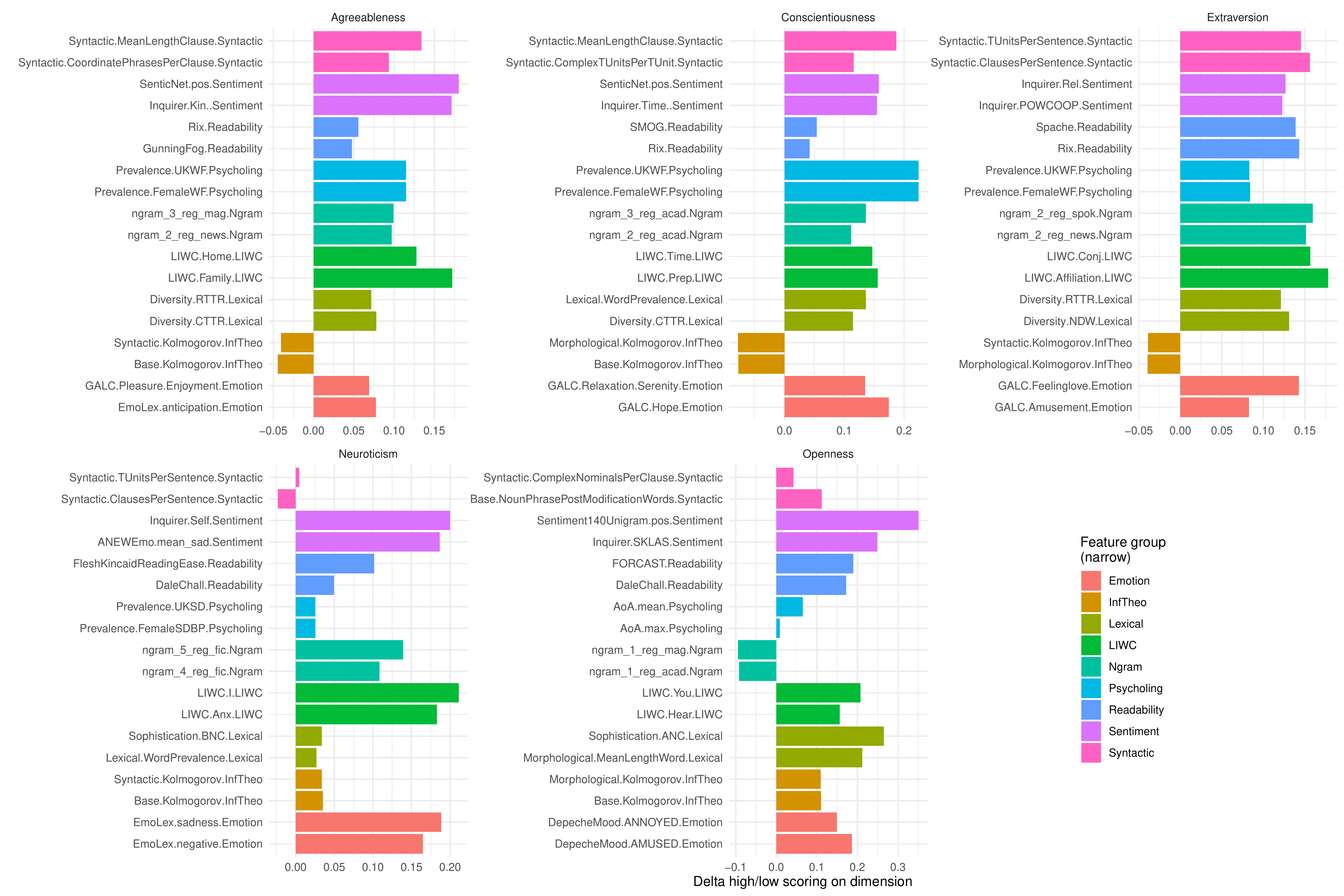}
    \caption{\textbf{Essays dataset:} Upper panel: Top 20 most characteristic features from each feature group by personality trait. Lower panel: Top 2 most characteristic features from each feature group by personality trait. Plotted scores represent the difference between the z-standardized mean scores of high- and low-scoring individuals on a given personality trait. Positive scores are characteristic of the high-scoring individuals on a given trait (e.g. individuals with high extraversion scores).}
    \label{fig:vis1}
\end{figure*}

\begin{figure*}
    \centering
    \includegraphics[width = 0.9\textwidth]{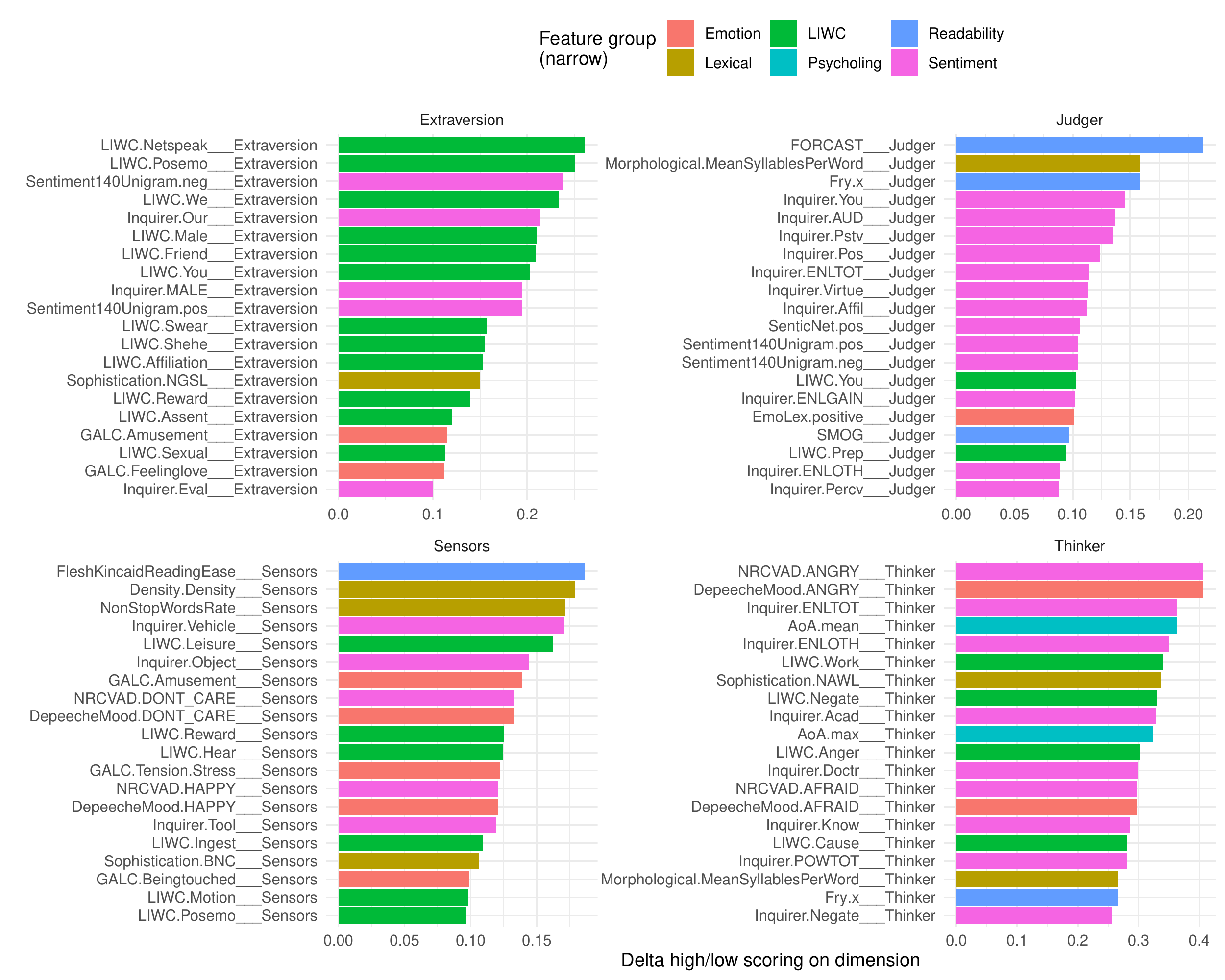}
    \includegraphics[width = 0.9\textwidth]{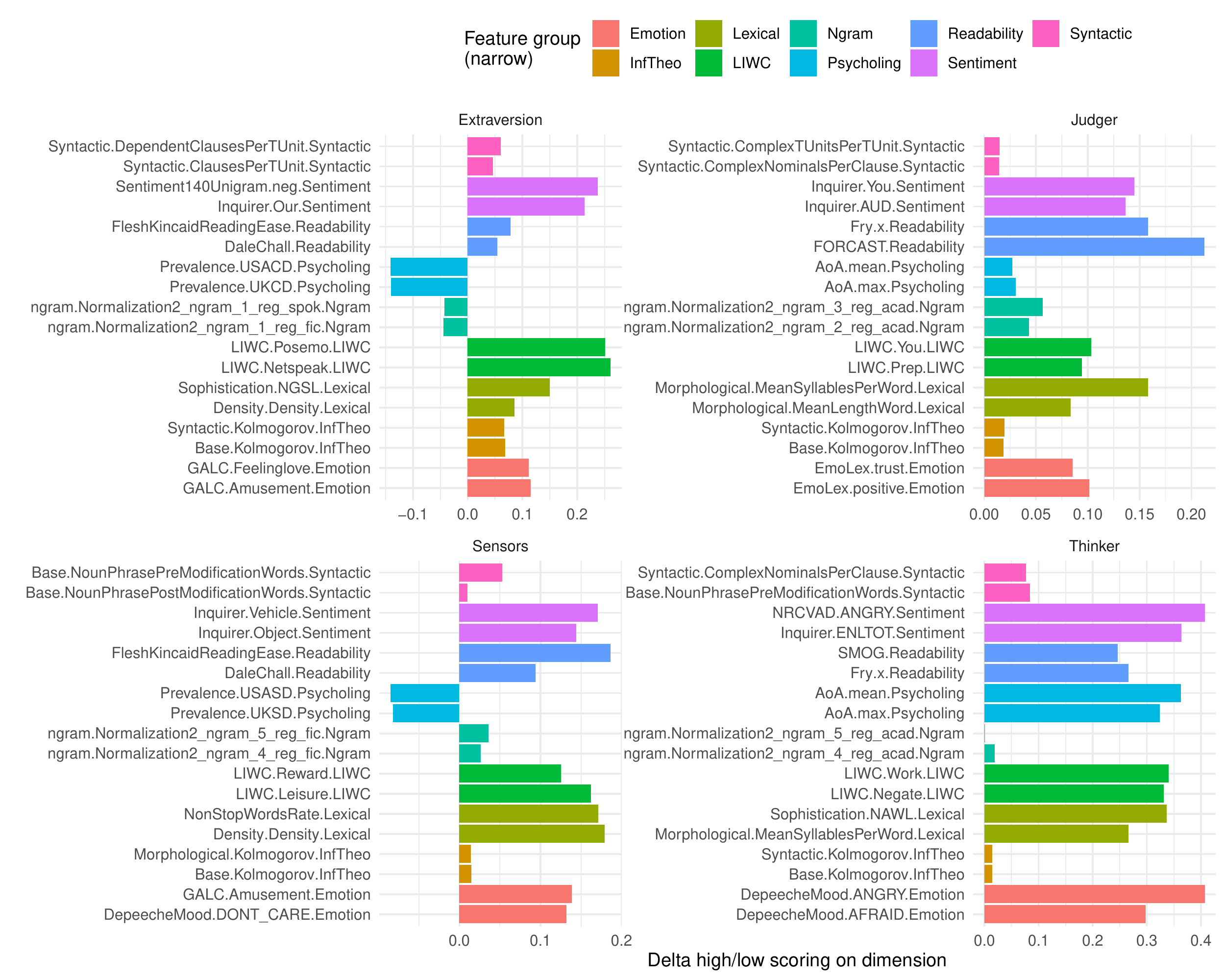}
    \caption{\textbf{MBTI Kaggle dataset:} Upper panel: Top 20 most characteristic features from each feature group by personality trait. Lower panel: Top 2 most characteristic features from each feature group by personality trait. Plotted scores represent the difference between the z-standardized mean scores of high- and low-scoring individuals on a given personality trait. Positive scores are characteristic of the high-scoring individuals on a given trait (e.g. individuals with high extraversion scores).}
    \label{fig:vis2}
\end{figure*}

\end{document}